%% file: neurips_2026.tex
\documentclass{article}
\usepackage[main,final]{neurips_2026}
\usepackage[utf8]{inputenc}
\usepackage[T1]{fontenc}
\usepackage{hyperref}
\usepackage{url}
\usepackage{booktabs}
\usepackage{amsfonts}
\usepackage{nicefrac}
\usepackage{microtype}
\usepackage{graphicx}
\usepackage{subcaption}
\usepackage{adjustbox}
\usepackage{amsmath,amssymb,mathtools,amsthm}
\usepackage{multirow}
\usepackage[table]{xcolor}
\usepackage{newunicodechar}
\newunicodechar{ć}{\'c}
\usepackage[capitalize,noabbrev]{cleveref}
\usepackage{algorithm}
\usepackage{algorithmic}
\hypersetup{pdftitle={DyJR: Preserving Local Policy Plasticity in Reinforcement Learning with Verifiable Rewards via Dynamic Jensen-Shannon Replay},pdfauthor={Long Li, Zhijian Zhou, Tianyi Wang, Weidi Xu, Zuming Huang, Wei Chu, Zhe Wang, Shirui Pan, Chao Qu, Yuan Qi},colorlinks=true,linkcolor=blue,citecolor=blue,urlcolor=blue}

\title{DyJR: Preserving Local Policy Plasticity in Reinforcement Learning with Verifiable Rewards via Dynamic Jensen-Shannon Replay}
\author{%
\textbf{Long Li}$^{1,*}$\quad \textbf{Zhijian Zhou}$^{2,3,*}$\quad \textbf{Tianyi Wang}$^{4,*}$\quad \textbf{Weidi Xu}$^{5}$\\
\textbf{Zuming Huang}$^{5}$\quad \textbf{Wei Chu}$^{5}$\quad \textbf{Zhe Wang}$^{1}$\quad \textbf{Shirui Pan}$^{1,\dagger}$\\
\textbf{Chao Qu}$^{2,6,\dagger}$\quad \textbf{Yuan Qi}$^{2,6}$\\[3pt]
\normalfont $^1$Griffith University, Australia\quad $^2$Fudan University, China\\
\normalfont $^3$Shanghai Innovation Institute, China\\
\normalfont $^4$Tencent Youtu Lab, China\\
\normalfont $^5$INFLY TECH, China\\
\normalfont $^6$Shanghai Academy of Artificial Intelligence for Science, China\\
\normalfont $^*$Equal contribution.\quad $^\dagger$Corresponding authors.\\
\normalfont Correspondence: \texttt{s.pan@griffith.edu.au}, \texttt{quchao@fudan.edu.cn}
}
\begin{document}
\maketitle
\begin{abstract}
While Reinforcement Learning (RL) enhances Large Language Model reasoning, on-policy algorithms like GRPO are sample-inefficient as they discard past rollouts. Existing experience replay methods address this by reusing accurate samples for direct policy updates, but this can incur additional storage costs and concentrate the policy on a narrow set of solutions. We argue that historical data should prioritize sustaining local policy plasticity rather than simply reinforcing accuracy. We define local policy plasticity as a persistent ability to remain backtrackable and switchable with respect to recent successful trajectories: throughout training, the policy maintains non-negligible support on recently reachable alternatives, so it can smoothly reallocate probability mass when needed without being locked into a single dominant path. To this end, we propose Dynamic Jensen-Shannon Replay (DyJR), a simple yet effective regularization framework using a dynamic reference distribution from recent trajectories. DyJR introduces two innovations: (1) A Time-Sensitive Dynamic Buffer that uses FIFO and adaptive sizing to retain only temporally proximal samples, synchronizing with model evolution; and (2) Jensen-Shannon Divergence Regularization, which replaces direct replay policy-gradient updates with a regularization signal that discourages premature over-commitment to a single Rank-1 trajectory. Experiments on mathematical reasoning, Text-to-SQL, and coding benchmarks demonstrate that DyJR improves over GRPO as well as baselines such as RLEP and Ex-GRPO, while maintaining training efficiency comparable to the original GRPO. Furthermore, Rank-$k$ token probability evolution and semantic-mode analyses show reduced concentration and greater retention of alternative continuations. Matched admission--objective comparisons and repeated runs provide additional evidence for the roles of DyJR's components.
\end{abstract}
\input{main_text/introduction}
\input{main_text/method}
\input{main_text/experiments}
\input{main_text/analysis}
\input{main_text/discussion}
\clearpage
\input{acknowledgments}
\bibliographystyle{plainnat}
\bibliography{example_paper}
\clearpage
\appendix
\input{appendix/limitations}
\input{appendix/method_details}
\input{appendix/original_experiments}

\input{appendix/rebuttal_additions}
\input{appendix/additional_analysis}
\clearpage
\input{checklist}
\end{document}

%% file: main_text/introduction.tex
\section{Introduction}
Reasoning capability is central to the development of large language models (LLMs)~\citep{gpt4,qwen3,zhu2025medeyes}. Models such as OpenAI o1 demonstrate that reinforcement learning (RL) can extend beyond preference alignment to improve reasoning through long chain-of-thought (CoT) exploration~\citep{o1,guo2025deepseek}. As ``long thinking'' with verifiable rewards becomes an increasingly important training paradigm~\citep{chen2025towards}, a key bottleneck remains: on-policy reinforcement learning with verifiable rewards (RLVR)~\citep{cai2025reinforcement,cai2026vi} discards expensive rollout data after the current update~\citep{deepseekmath,zhan2025exgrpo}. This limits reuse of past successes and increases the cost of scaling RL to long reasoning trajectories~\citep{fu2025areallargescaleasynchronousreinforcement,zhou2026look}. Experience replay~\citep{rolnick2019experience,mnih2013playing,lillicrap2015continuous,liu2026automated} offers a natural way to improve sample efficiency. Methods such as ReMix~\citep{liang2025squeeze}, RePO~\citep{li2025repo}, Ex-GRPO~\citep{zhan2025exgrpo}, and RLEP~\citep{zhang2025rlep} reuse historical trajectories for additional policy updates. In this data-augmentation view, replay contributes a second reward-optimization term alongside the online objective, schematically $\mathcal J(\theta)=\mathcal J_{\mathrm{on}}(\theta)+\alpha\mathcal J_{\mathrm{exp}}(\theta)$. Although admission rules and importance corrections differ across methods, the common purpose is to learn again from previously generated trajectories. We ask whether successful historical data can also serve a different role: maintaining access to useful alternatives as the current policy evolves.

Two challenges motivate this question. First, repeatedly reinforcing a restricted collection of successful trajectories can concentrate the policy on specific solution paths~\citep{zhu2025surprising,peng2025simko}, reducing the alternatives available for further exploration~\citep{wang20258020rulehighentropyminority,yue2025does,cheng2025reasoningexplorationentropyperspective}. High reward alone does not reveal whether the policy still supports several ways of reaching a solution. Second, retaining a large history increases storage and mixes trajectories generated at different stages of training. Our training diagnostics reveal a \emph{rapid transition window}: entropy decreases sharply during the early steps before the policy enters a more concentrated regime. A replay strategy must therefore provide enough successful trajectories during this transition while keeping its later reference relevant to the current policy. Based on these observations, we study experience replay as a regularization mechanism for sustaining \emph{local policy plasticity}. By this term, we mean the persistent ability to remain backtrackable and switchable with respect to recent successful trajectories: the policy retains non-negligible support on nearby alternatives, so it can reallocate probability mass instead of becoming locked into a single dominant continuation. This does not require preserving every behavior encountered throughout training. Nor does a rise in token entropy alone establish that the retained alternatives correspond to useful reasoning strategies. We examine token dynamics together with separate semantic-mode measurements to assess this interpretation.

We propose \emph{Dynamic Jensen-Shannon Replay} (DyJR), with two complementary innovations. Regarding \textbf{data construction}, a confidence-stratified admission rule selects successful trajectories for a first-in-first-out (FIFO) buffer. A higher admission rate populates the buffer during the initial transition, followed by a lower rate as training stabilizes. A short maximum-age window retains temporally adjacent samples and bounds the history used for regularization. Regarding \textbf{data utilization}, DyJR replaces additional replay reward-gradient updates with a Jensen-Shannon (JS) regularizer. Recent successful trajectories, together with their cached generation-time log-probabilities, provide a dynamic anchor that discourages those continuations from losing support. GRPO continues to learn from current rollouts, while replay helps preserve recently reachable alternatives. The distinction between construction and utilization is essential. Under a matched replay budget and freshness constraint, confidence-based admission improves the six-benchmark average by 2.5 points for off-policy replay and by 3.7 points for JS regularization. With this admission rule fixed, JS adds a further 1.5 points over off-policy replay. These controls support the complementary roles of admission and regularization rather than attributing the full gain to the divergence alone.

Our main contributions are threefold. \textbf{(1) Replay as plasticity regularization.} We introduce a replay framework that uses recent successful trajectories as a distributional reference, separating maintenance of local successful support from direct reinforcement of replayed rewards. \textbf{(2) Temporally local reference construction.} We combine confidence-stratified admission, a larger early admission rate, and a short FIFO window. The resulting reference adapts to policy evolution while avoiding storage of the full training history. \textbf{(3) Empirical evaluation and mechanism analysis.} We evaluate mathematics, Text-to-SQL, and coding with Qwen and Llama models, including repeated-run replay comparisons and matched mechanism controls. Rank-$k$ token dynamics, sampling-budget scaling, and semantic-mode analyses provide complementary evidence about how data construction and utilization affect training.

%% file: main_text/method.tex
\begin{figure}[t]
\centering
\includegraphics[width=\linewidth]{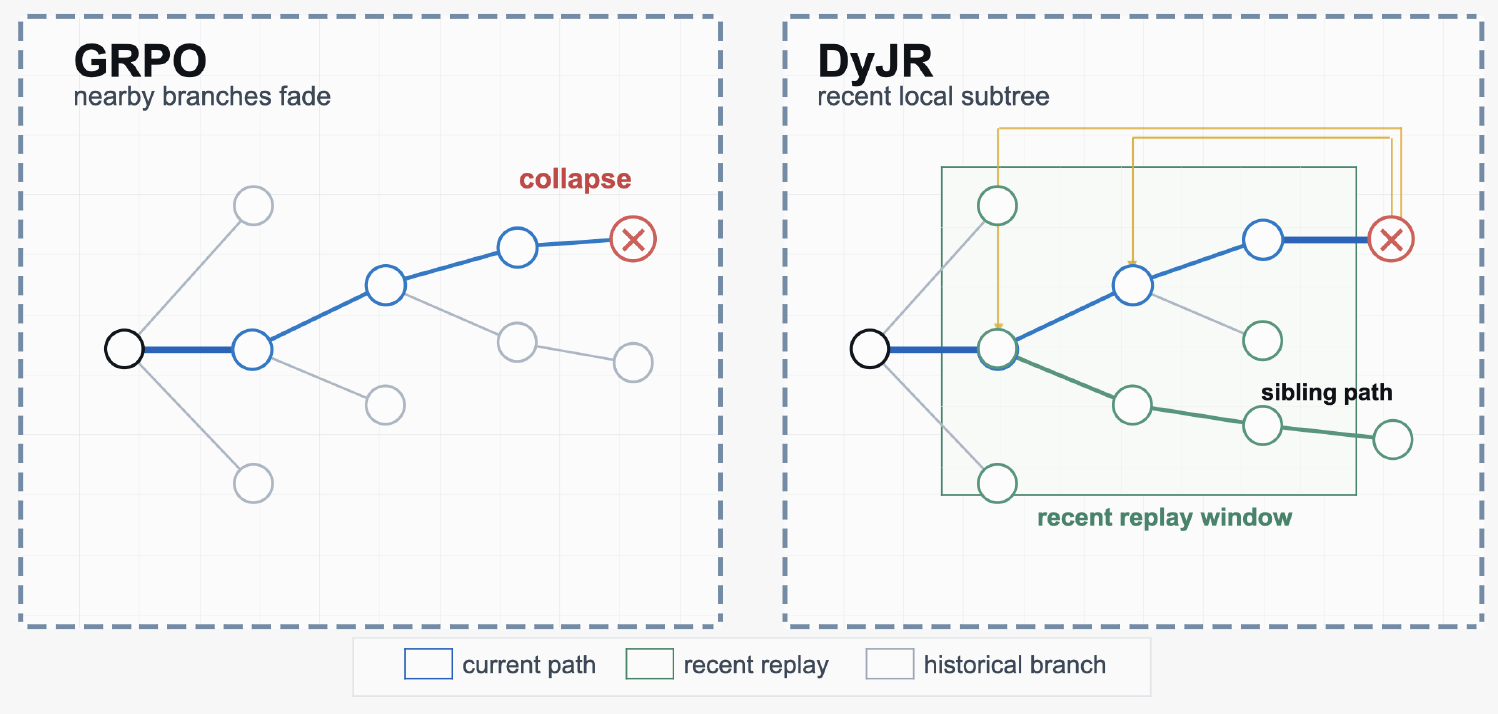}
\caption{Conceptual illustration of local replay. DyJR retains recent successful branches as regularization targets when current sampling becomes concentrated. The tree illustrates accessible alternatives; the algorithm does not perform explicit tree search or a backtracking procedure.}
\label{fig:dyjr_method_tree_replay}
\end{figure}

\section{Preliminaries}
\label{sec:preliminaries}
\paragraph{RLVR and GRPO.}
A policy $\pi_\theta$ generates a response $y$ to prompt $x$ and receives a verifiable reward $r(x,y)$. GRPO~\citep{deepseekmath} samples $G$ responses per prompt from $\pi_{\theta_{\mathrm{old}}}$ and computes group-normalized advantages $\widehat A_i=(r(x,y_i)-\mu_x)/(\sigma_x+\varepsilon)$, where $\mu_x$ and $\sigma_x$ are the group reward mean and standard deviation and $\varepsilon$ is a numerical stabilizer. At token state $s_{i,j}=(x,y_i^{(<j)})$, let $\rho_{i,j}=\pi_\theta(y_i^{(j)}\mid s_{i,j})/\pi_{\theta_{\mathrm{old}}}(y_i^{(j)}\mid s_{i,j})$. The clipped token loss is
\begin{equation}
\ell_{i,j}^{\mathrm{GRPO}}=-\min\!\left[\rho_{i,j}\widehat A_i,\ \operatorname{clip}(\rho_{i,j},1-\epsilon_{\mathrm{low}},1+\epsilon_{\mathrm{high}})\widehat A_i\right].
\label{eq:prelim-grpo}
\end{equation}
We average over valid response tokens and omit the fixed-reference KL penalty.

\paragraph{Experience replay.}
Historical trajectories come from earlier policies. Direct replay uses them for reward-driven updates, whereas DyJR uses recent successes as regularization targets alongside the online GRPO loss. The online ratio uses this step's rollout policy; replay instead uses each trajectory's historical generating policy.

\paragraph{Jensen-Shannon divergence.}
For distributions $P$ and $Q$, let $R=(P+Q)/2$. Jensen-Shannon divergence is $D_{\mathrm{JS}}(P\|Q)=\tfrac12 D_{\mathrm{KL}}(P\|R)+\tfrac12 D_{\mathrm{KL}}(Q\|R)$. It is symmetric and bounded by $\ln2$ with natural logarithms. DyJR uses a sampled surrogate with cached historical probabilities, detailed in Appendix~\ref{app:divergence-background}.

\section{Method}
\label{sec:method}
DyJR combines a \textbf{Dynamic Replay Buffer} with \textbf{JS Regularization} to preserve recently successful alternatives as the policy evolves. Its two components address data construction and data utilization: the buffer selects a temporally local reference, and the regularizer uses that reference as a flexible anchor alongside the online GRPO objective. Figure~\ref{fig:dyjr_method_tree_replay} illustrates the mechanism; the full procedure appears in Algorithm~\ref{alg:dyjr}.

\paragraph{Local Policy Plasticity.}
We use \emph{local policy plasticity} to denote a \emph{persistent} ability to remain \emph{backtrackable and switchable} with respect to recent successful trajectories. The aim is to retain alternatives throughout training, so probability mass can shift when the current dominant continuation becomes over-committed. At a prefix state $s$, let $a_k$ be the token with the $k$-th highest probability under the current policy, so $a_1$ is its dominant continuation. Let $q_t(\cdot\mid s)$ denote the conceptual reference induced by recent successful trajectories. Our objective is to retain non-negligible support on replay-supported alternatives:
\begin{equation}
q_t(a_k\mid s)>0,\ k\geq2 \quad\Longrightarrow\quad
\pi_{\theta_t}(a_k\mid s)\not\approx0.
\label{eq:plasticity}
\end{equation}
This notion concerns local adjustability: it neither requires coverage of every mode encountered during training nor merely asks the policy to change slowly. Equation~\ref{eq:plasticity} describes the intended behavior rather than a guarantee. Because token alternatives need not represent different reasoning modes, we assess both token-level diagnostics and semantic-mode analyses in Section~\ref{sec:semantic-analysis}.

\subsection{Dynamic Reference Construction}
A buffer entry stores a prompt, response, generation step $\tau$, and the sampled response tokens' pre-update generating-policy log-probabilities $\ell_{\mathrm{old}}$. Before the update at step $t$, the usable buffer contains only previously admitted successful trajectories:
\begin{equation}
\mathcal S_t=\{(x,y,\ell_{\mathrm{old}},\tau):\text{admitted at }\tau<t,\ r(x,y)=1,\ t-\tau\leq M\}.
\label{eq:buffer}
\end{equation}
FIFO eviction enforces the maximum age $M$. The goal is to preserve \emph{locally reachable alternatives} within the evolving policy neighborhood. Explicit semantic selection is difficult for long-chain reasoning, where surface-form similarity can misrepresent solution semantics; recency provides an inexpensive, task-agnostic proxy for policy relevance. Small clipped updates motivate this local view but do not guarantee bounded drift. A short window limits the exposure to stale anchors, with its practical effect assessed by the staleness ablation.

For each prompt $x$, define empirical confidence $C(x)$ as the number of correct responses among $G$ current rollouts. We visit confidence levels from $G$ down to $1$, admitting only correct trajectories until a target fraction $\eta_t$ of the $N$ online rollouts is reached:
\begin{equation}
\mathcal P_t\leftarrow\mathcal P_t\cup
\{(x,y,\ell_{\mathrm{old}},t):C(x)=k,\ r(x,y)=1\},\quad k=G,G-1,\ldots,1.
\label{eq:selection}
\end{equation}
The target is a capacity budget, not a requirement to admit incorrect responses when successes are scarce. This high-to-low strategy first secures successful trajectories from high-confidence prompt groups, then relaxes the confidence criterion when these do not fill the budget. Compared with admitting only fully successful groups, it can therefore include rare successful trajectories from harder prompts. This deliberately selected reference remains biased toward successful responses; it is not an unbiased estimate of the rollout distribution. We use $\eta_t=20\%$ for the first 20 training steps and $5\%$ thereafter, with $M=8$. Early RLVR training can sharply concentrate the policy distribution. The larger initial rate helps populate the buffer during this rapid transition, retaining successful early alternatives before they become difficult to sample. The admission rates and $M$ were selected in initial pilots and reused for the 1.7B model without retuning. Section~\ref{sec:admission-results} distinguishes the evidence for early buffer population from the storage benefit of later contraction.

\subsection{JS Regularization Implementation}
We implement a sampled, token-level JS-style objective~\citep{Wang2023BeyondRK}. For replay trajectory $s$ and response position $j$, define
\begin{align}
u_s^{(j)}&=\exp\!\left[\log\pi_\theta(y_s^{(j)}\mid x_s,y_s^{(<j)})-\ell_{\mathrm{old},s}^{(j)}\right],\label{eq:js_ratio}\\
f(u)&=u\log u-(u+1)\log\frac{u+1}{2},\label{eq:js-generator}\\
\mathcal L_{\mathrm{JS}}(\theta)&=\frac{\sum_{s\in B_{\mathrm{replay}}}\sum_j m_s^{(j)}f(u_s^{(j)})}{\sum_{s\in B_{\mathrm{replay}}}\sum_j m_s^{(j)}}.
\label{eq:js_loss}
\end{align}
Here $B_{\mathrm{replay}}$ denotes replay rows in the current optimization micro-batch, and $m_s^{(j)}$ is its valid-response-token mask. Aggregation is token-mean within this masked batch, with zero replay contribution if it contains no valid replay tokens; cached log-probabilities remain fixed. This avoids re-evaluating the historical generating policy. The regularizer uses the generating policy associated with each trajectory; it does not explicitly construct or evaluate the full mixture density $q_t$.

The displayed generator gives $\mathbb E_{z\sim Q}f(P(z)/Q(z))=2D_{\mathrm{JS}}(P\|Q)$ under exact reference sampling, with its normalization absorbed into the coefficient. Success filtering and cached per-policy denominators make this a surrogate rather than an unbiased mixture-divergence estimate. Bounded JS does not bound individual sample terms or gradients; Appendix~\ref{app:divergence-background} details these distinctions.

\subsection{Joint Optimization Objective}
GRPO encourages higher-reward solutions while replay regularization anchors the policy to recent successful alternatives. The joint objective is
\begin{equation}
\mathcal L_{\mathrm{total}}=\mathcal L_{\mathrm{GRPO}}+\alpha_{\mathrm{JS}}\mathcal L_{\mathrm{JS}}.
\label{eq:total-loss}
\end{equation}
We use $\alpha_{\mathrm{JS}}=0.05$. Our tuning rule keeps the online policy loss and weighted JS loss at roughly a $10{:}1$ scale. Starting with a short recency window, we adjust $\alpha_{\mathrm{JS}}$ so the anchor remains about one order of magnitude weaker than the online objective. This balances an ineffective anchor against over-constraining policy updates; it is a loss-scale heuristic, not a fixed relationship between gradients.

Each step generates 4,096 online trajectories from 512 prompts with eight responses each. If the buffer contains at least 512 historical trajectories, we sample 512 uniformly without replacement once and hold them fixed throughout the step; otherwise replay is skipped. One epoch over the 4,608 shuffled rows, with mini-batch size 128, gives 36 optimizer steps. Online and replay rows contribute to GRPO and JS, respectively. Current successful trajectories enter the buffer only after optimization, retaining their rollout-time log-probabilities. Algorithm~\ref{alg:dyjr} gives the complete ordering and batching details.

%% file: main_text/experiments.tex
\section{Experiments}
\label{sec:experiments}
\subsection{Tasks, Baselines, and Evaluation}
We evaluate three task families. \textbf{Mathematics:} Qwen3-4B-Base is trained on Reinforce-Ada-Hard, a 14k-problem subset of OpenR1-Math-220k~\citep{xiong2025reinforce}, with responses up to 6,144 tokens. We evaluate AIME25, AMC23, Beyond AIME, BRUMO25, HMMT25, and Minerva, and also test Qwen3-1.7B-Base under sparser rewards. \textbf{Text-to-SQL:} Llama-3.1-8B-Instruct is trained on BIRD and evaluated on BIRD and Spider~\citep{li2024bird,yu2018spider}. \textbf{Coding:} Qwen2.5-7B-Instruct is trained on CodeContests and evaluated on LiveCodeBench-V2 and held-out CodeContests. Appendix~\ref{expset} gives configurations and metrics.

Baselines are GRPO without the reference-KL penalty, DAPO~\citep{dapo}, DPH-RL~\citep{li2025choice}, RLEP~\citep{zhang2025rlep}, and Ex-GRPO~\citep{zhan2025exgrpo}. All are reproduced in the same VERL training stack, retaining their method-specific replay settings. We use an upper clipping parameter of 0.28 throughout. The main mathematical runs use AdamW with a learning rate of $2\times10^{-5}$, weight decay 0.1, and gradient clipping at 1.0. Each run uses 32 A800 GPUs for approximately 20 hours. Unless stated otherwise, ablations change the specified component of DyJR under the same training budget. Checkpoints are evaluated every 80 steps on the six mathematical benchmarks, without a separate validation split. Selecting the highest six-benchmark average yields the same final checkpoint, step 240, for every method and run in the main comparison. Thus, all reported results use a common endpoint. GRPO, DyJR, RLEP, and Ex-GRPO have three independent runs in the 4B comparison; other rows and ablations are single runs unless specified. Appendix~\ref{app:rebuttal-repeated-runs} reports the run-level results and sample standard deviations across three runs.

\subsection{Mathematical Reasoning}
\begin{table}[t]
\centering
\caption{Mathematical reasoning accuracy (\%). Minerva uses Mean@16; the other benchmarks use Mean@256. The 4B GRPO, DyJR, RLEP, and Ex-GRPO rows show three-run means; other rows are single runs. Run-level variability is reported in Appendix~\ref{app:rebuttal-repeated-runs}. Bold marks the best main-method result within each model size.}
\label{tab:math_results}
\small
\setlength{\tabcolsep}{3pt}
\renewcommand{\arraystretch}{1.18}
\begin{tabular*}{\linewidth}{@{\extracolsep{\fill}}lrrrrrrr@{}}
\toprule
Method & AIME25 & AMC23 & \shortstack{Beyond\\AIME} & BRUMO25 & HMMT25 & Minerva & Avg \\
\midrule
\multicolumn{8}{l}{\textbf{Qwen3-4B-Base}}\\
Base model & 8.2 & 51.5 & 4.3 & 16.6 & 1.9 & 29.0 & 18.6\\
GRPO & $19.1$ & $64.6$ & $10.7$ & $29.6$ & $9.8$ & $45.2$ & 29.8\\
DAPO &21.5&65.5&10.6&30.4&9.2&47.8&30.8\\
DPH-RL &22.8&65.2&10.8&30.1&10.5&48.1&31.3\\
RLEP & $22.6$ & $70.9$ & $10.9$ & $28.7$ & $10.7$ & $47.3$ & $31.8$\\
Ex-GRPO & $22.2$ & $69.9$ & $11.4$ & $33.0$ & $10.0$ & $49.2$ & $32.6$\\
\rowcolor{black!4}
DyJR & $\textbf{23.1}$ & $\textbf{72.0}$ & $\textbf{12.7}$ & $\textbf{33.4}$ & $\textbf{12.7}$ & $\textbf{50.4}$ & $\textbf{34.1}$\\
\midrule
\multicolumn{8}{l}{\textbf{Qwen3-1.7B-Base (sparse rewards)}}\\
GRPO &8.5&47.8&3.1&11.7&\textbf{1.6}&31.3&17.3\\
DyJR &\textbf{10.1}&\textbf{53.8}&\textbf{3.9}&\textbf{14.4}&\textbf{1.6}&\textbf{33.4}&\textbf{19.5}\\
\bottomrule
\end{tabular*}
\end{table}

Table~\ref{tab:math_results} shows that DyJR improves the six-benchmark average from GRPO's 29.8 to 34.1, a gain of 4.3 percentage points. It also exceeds Ex-GRPO (32.6) and RLEP (31.8). Appendix~\ref{app:rebuttal-repeated-runs} gives benchmark-level variability and all newly repeated baseline runs. The remaining single-run ablations do not establish statistical significance. In the weaker Qwen3-1.7B setting, DyJR improves the average from 17.3 to 19.5. We reuse $M=8$, the 20\%/5\% admission schedule, the replay-to-online ratio, and the loss-scale heuristic rather than retuning these defaults. This demonstrates transfer between the tested sizes without establishing universally optimal defaults.

\subsection{Separating Admission from Replay Utilization}
\label{sec:admission-results}
\begin{table}[t]
\centering
\caption{Matched admission--objective comparison at $M=8$. Values are six-benchmark averages (\%). The training budget, replay batch size, schedule, and freshness constraint are fixed. PG+IS denotes off-policy policy gradients with importance ratios.}
\label{tab:admission-main}
\renewcommand{\arraystretch}{1.18}
\setlength{\tabcolsep}{12pt}
\begin{tabular}{lrr}
\toprule
Admission rule & PG+IS & JS regularization\\
\midrule
Perfect-only &30.1&30.4\\
Confidence-based &32.6&\textbf{34.1}\\
\bottomrule
\end{tabular}
\end{table}
The matched comparison in Table~\ref{tab:admission-main} isolates two complementary choices. Confidence-based admission improves PG+IS by 2.5 points and JS by 3.7 points. With this admission rule fixed, JS improves over PG+IS by 1.5 points; under perfect-only admission, the difference is only 0.3 points. Adequate buffer population is therefore an important prerequisite in the sparse-success setting. The auxiliary controls are single runs; their comparison with the reported DyJR mean supports a conditional benefit of JS, rather than a decomposition that assigns the entire DyJR--GRPO gap to the regularizer.

The admission schedule also separates early availability from later storage cost. Constant 20\% admission yields 34.2, close to 34.1 for the 20\%$\rightarrow$5\% schedule. Reducing initial admission to 10\% or 5\% lowers the average to 32.4 or 31.3. Thus, later contraction reduces admission and capacity without an observed performance loss in this comparison, while insufficient early population is harmful. It does not establish that late trajectories are intrinsically less valuable (Appendix~\ref{app:admission-schedule}).

\subsection{Reference, Divergence, and Entropy Controls}
\label{sec:reference-controls}
\begin{table}[t]
\centering
\caption{Ablations of the replay reference, divergence, regularization coefficient, and recency window. Values are mathematical accuracy (\%); other settings follow DyJR. The default is $\alpha_{\mathrm{JS}}=0.05$ and $M=8$. Controls are single runs; the DyJR row shows its three-run mean.}
\label{tab:math_controls}
\small
\setlength{\tabcolsep}{3pt}
\renewcommand{\arraystretch}{1.18}
\begin{tabular*}{\linewidth}{@{\extracolsep{\fill}}lrrrrrrr@{}}
\toprule
Setting & AIME25 & AMC23 & \shortstack{Beyond\\AIME} & BRUMO25 & HMMT25 & Minerva & Avg \\
\midrule
\rowcolor{black!4}
DyJR (default) &23.1&72.0&12.7&33.4&12.7&50.4&\textbf{34.1}\\
\midrule
\multicolumn{8}{l}{\textit{Divergence and reference construction}}\\
Forward KL &22.8&70.7&11.3&32.6&10.5&47.3&32.5\\
Static Ref + JS &22.7&66.3&11.2&29.9&10.7&49.4&31.7\\
Reverse-KL (recent) &18.7&64.3&10.1&28.8&10.1&44.3&29.4\\
\midrule
\multicolumn{8}{l}{\textit{Regularization coefficient and recency window}}\\
$\alpha_{\mathrm{JS}}=0.01$ &22.5&69.3&11.5&31.0&12.5&49.5&32.7\\
$\alpha_{\mathrm{JS}}=0.2$ &21.4&66.7&10.1&29.0&10.9&47.4&30.9\\
$M=16$ &23.6&69.3&11.3&31.0&13.6&48.8&32.9\\
$M=32$ &22.0&68.9&10.9&30.2&10.3&47.7&31.7\\
$M=64$ &21.3&67.9&10.5&29.3&9.7&46.9&30.9\\
\bottomrule
\end{tabular*}
\end{table}

\begin{table}[t]
\centering
\caption{Cross-domain results (\%). SQL columns report execution-based Pass@1 and Pass@16. Coding columns report Pass@16 under unit-test verification. These are single-run comparisons. Bold marks the best value in each column.}
\label{tab:cross-domain}
\small
\setlength{\tabcolsep}{4pt}
\renewcommand{\arraystretch}{1.18}
\begin{tabular*}{\linewidth}{@{\extracolsep{\fill}}lrrrrrr@{}}
\toprule
&\multicolumn{2}{c}{BIRD}&\multicolumn{2}{c}{Spider}&\multicolumn{2}{c}{Coding (Pass@16)}\\
\cmidrule(lr){2-3}\cmidrule(lr){4-5}\cmidrule(lr){6-7}
Method&Pass@1&Pass@16&Pass@1&Pass@16&\shortstack{LiveCodeBench\\V2}&CodeContests\\
\midrule
GRPO&59.4&68.4&72.5&80.3&30.9&25.5\\
DAPO&60.1&69.2&71.5&79.4&30.7&26.4\\
DPH-RL&\textbf{62.8}&72.4&76.0&84.1&31.2&26.7\\
\rowcolor{black!4}
DyJR&62.7&\textbf{72.9}&\textbf{77.5}&\textbf{87.3}&\textbf{31.8}&\textbf{27.2}\\
\bottomrule
\end{tabular*}
\end{table}

Table~\ref{tab:math_controls} compares reference construction and divergence choices. Holding the replay budget, admission rule, and coefficient fixed, Static Ref + JS obtains 31.7 versus 34.1 for DyJR. DPH-RL, another JS-based method, obtains 31.3. These comparisons emphasize reference construction rather than identifying the divergence alone as the source of improvement. Reverse-KL to recent historical anchors obtains 29.4, so conservative anchoring alone is insufficient in this setting. Replacing JS with forward KL yields 32.5. Together these results favor the tested combination of a recent reference and JS-style utilization, without constituting a general theorem about either divergence.

Standard token-level entropy regularization yields 29.7, compared with 29.8 for GRPO and 34.1 for DyJR. The separately evaluated dynamic entropy bonus obtains 30.3; these are distinct controls (Appendix~\ref{app:rebuttal-entropy}). An undirected entropy term need not preferentially preserve successful alternative continuations. For recency, increasing $M$ from 8 to 16, 32, or 64 reduces the average to 32.9, 31.7, or 30.9. The effect of a longer window also depends on admission: with perfect-only selection, increasing $M$ from 8 to 32 raises the average from 30.4 to 31.3, whereas confidence-based selection falls from 34.1 to 31.7 under the same change (Appendix~\ref{app:selection-sparse-rewards}). A longer history can alleviate scarcity under a restrictive admission rule, while older anchors may be less useful once enough recent successes are available. This interaction motivates choosing admission and freshness together; neither a short window nor a large buffer is uniformly preferable across these controls. For regularization strength, coefficients 0.01 and 0.2 yield 32.7 and 30.9. The latter is an intentionally strong stress-test setting, still above GRPO, rather than evidence of a sharp instability boundary.

\subsection{Transfer to SQL and Coding}

DyJR improves over GRPO on both SQL benchmarks and both coding benchmarks (Table~\ref{tab:cross-domain}). On BIRD and Spider, Pass@1 increases by 3.3 and 5.0 points, and Pass@16 by 4.5 and 7.0 points. DPH-RL is slightly higher on BIRD Pass@1 (62.8 versus 62.7), while DyJR has the highest values in the other reported columns. Coding gains over GRPO are 0.9 and 1.7 points. These results extend the evaluation beyond competition mathematics; their single-run nature limits conclusions about smaller gaps.

%% file: main_text/analysis.tex
\section{Analysis}
\label{sec:analysis}
\subsection{Token-Level Training Dynamics}
\label{sec:plasticity_evolution}
Following prior token-probability diagnostics~\citep{peng2025simko}, we track a top-$k$-normalized entropy and the average probability of each ranked token. For top-$k$ log-probabilities $\ell_i$, define $\widetilde p_i=\exp(\ell_i)/\sum_{j=1}^k\exp(\ell_j)$ and $H_{\mathrm{top}\text{-}k}=-\sum_i\widetilde p_i\log\widetilde p_i$. This is entropy within the retained token set, not full-vocabulary entropy. The average rank-$k$ probability is $\overline P_{\mathrm{rank}\text{-}k}=N^{-1}\sum_{n=1}^N\exp(\ell_{\mathrm{rank}\text{-}k}^{(n)})$.

\begin{figure}[t]
\centering
\includegraphics[width=\linewidth]{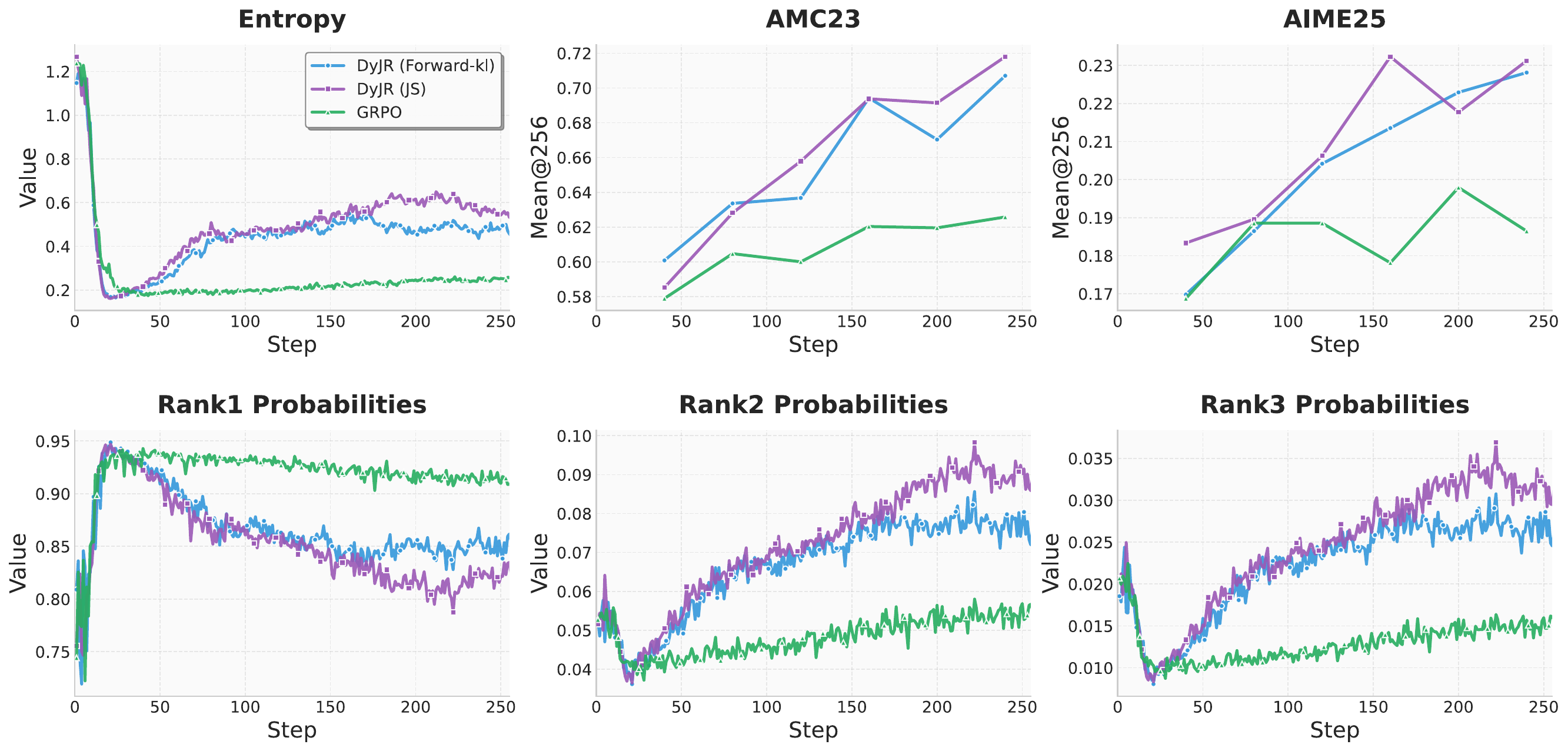}
\caption{Qwen3-4B-Base training dynamics: top-$k$-normalized entropy, average ranked-token probabilities, and evaluation accuracy. Token concentration is an indirect diagnostic of the local-plasticity interpretation, rather than a measurement of semantic modes.}
\label{fig:training_divergence}
\end{figure}
In Figure~\ref{fig:training_divergence}, GRPO rapidly concentrates probability on rank-1 tokens and remains highly concentrated. DyJR also adapts sharply at the start but later assigns more mass to rank-2 and rank-3 alternatives. The forward-KL variant lies between the two in several diagnostics and remains below JS in the reported accuracy comparisons. These patterns are consistent with reduced token-level concentration under DyJR. They do not establish that each additional token alternative is useful, or that token concentration alone causes a performance plateau.

\subsection{From RLEP to DyJR}
\begin{figure}[t]
\centering
\includegraphics[width=\linewidth]{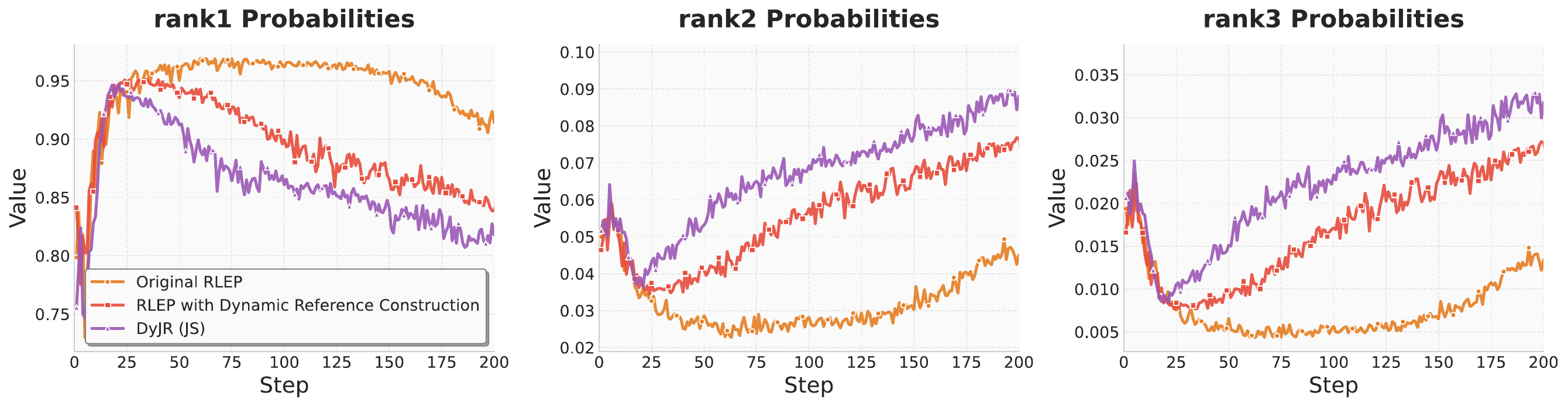}
\caption{Rank-$k$ probability trends for RLEP, RLEP with dynamic reference construction, and complete DyJR.}
\label{fig:rlep2dyjr}
\end{figure}
Figure~\ref{fig:rlep2dyjr} compares original RLEP, a variant that replaces its sampling mechanism with DyJR's dynamic reference construction while retaining replay policy updates, and complete DyJR. Dynamic construction reduces rank-1 concentration, and JS utilization further shifts mass toward alternatives. The construction variant changes more than a single scalar hyperparameter; the matched admission--objective experiment in Table~\ref{tab:admission-main} provides a more direct comparison of those two choices. We interpret the figure as a descriptive account of their training dynamics rather than a causal isolation of JS.

\subsection{Semantic Modes Beyond Token Statistics}
\label{sec:semantic-analysis}
\begin{table}[t]
\centering
\small
\caption{Mathematical strategy diversity on the same 50 problems, with 16 complete derivations per problem and model. Columns 2--6 give the percentage of problems. Modes include coherent incorrect strategies; the final column is the reported mean number of modes per problem.}
\label{tab:rebuttal-math-modes}
\setlength{\tabcolsep}{4pt}
\renewcommand{\arraystretch}{1.18}
\begin{tabular*}{\linewidth}{@{\extracolsep{\fill}}lrrrrrr@{}}
\toprule
Model & 1 mode & 2 modes & 3 modes & $\geq4$ modes & $\geq2$ modes & \shortstack{Mean\\modes} \\
\midrule
Qwen3-4B-Base & 22\% & 24\% & 34\% & 20\% & 78\% & 2.6 \\
GRPO & 88\% & 4\% & 6\% & 2\% & 12\% & 1.3 \\
\rowcolor{black!4}
DyJR & 42\% & 18\% & 24\% & 16\% & 58\% & 2.2 \\
\bottomrule
\end{tabular*}
\end{table}
We supplement these proxies with two protocol-specific analyses. In controlled NL2SQL, supervised fine-tuning constructs five solution families with identifying prefixes. With 32 responses per question, 95\% of questions yield only one family after GRPO, whereas 89\% yield at least two after DyJR. Prefixes label predefined protocols; they do not independently verify the meaning or correctness of an unconstrained response.

Table~\ref{tab:rebuttal-math-modes} reports the mathematical strategy distribution. We fix 50 problems and sample 16 complete derivations per problem and model. GPT-5.6-sol clusters executed mathematical strategies using all responses and their correctness labels, merging surface variation and retaining coherent incorrect attempts. Differences in wording, calculation order, arithmetic errors, and minor reparameterizations are merged when the core strategy is unchanged; responses without an identifiable strategy are excluded. Thus, the count measures distinct mathematical approaches rather than surface forms, and does not require every identified approach to contain a correct solution. The proportion of problems with at least two modes is 78\% for Qwen3-4B-Base, 12\% for GRPO, and 58\% for DyJR; mean mode counts are 2.6, 1.3, and 2.2. DyJR therefore retains more observed strategy diversity than GRPO on this set, while remaining below the base model. The small sample, judge dependence, and inclusion of incorrect strategies limit generalization. Appendix~\ref{app:rebuttal-sql-modes}--\ref{app:rebuttal-math-modes} gives the complete protocols and SQL distribution.

\subsection{Sampling Budget and Computational Cost}
\begin{figure}[t]
    \centering
    \captionsetup[subfigure]{justification=centering}

    \begin{subfigure}[b]{0.31\textwidth} 
        \centering
        \includegraphics[width=\textwidth, keepaspectratio]{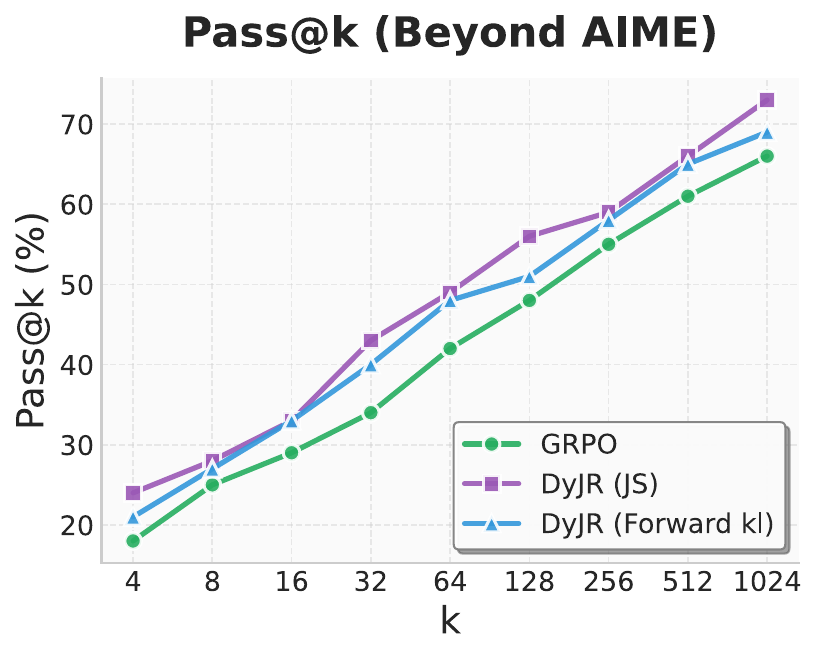}
        \caption{}
        \label{fig:sub_a}
    \end{subfigure}
    \hfill
    \begin{subfigure}[b]{0.33\textwidth}
        \centering
        \includegraphics[width=\textwidth, keepaspectratio]{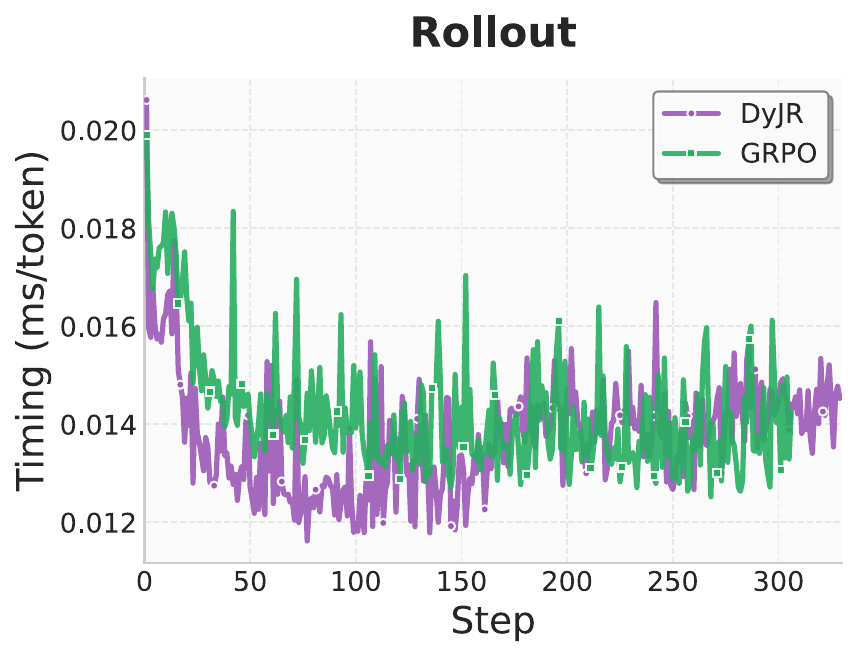}
        \caption{}
        \label{fig:sub_b}
    \end{subfigure}
    \hfill
    \begin{subfigure}[b]{0.33\textwidth}
        \centering
        \includegraphics[width=\textwidth, keepaspectratio]{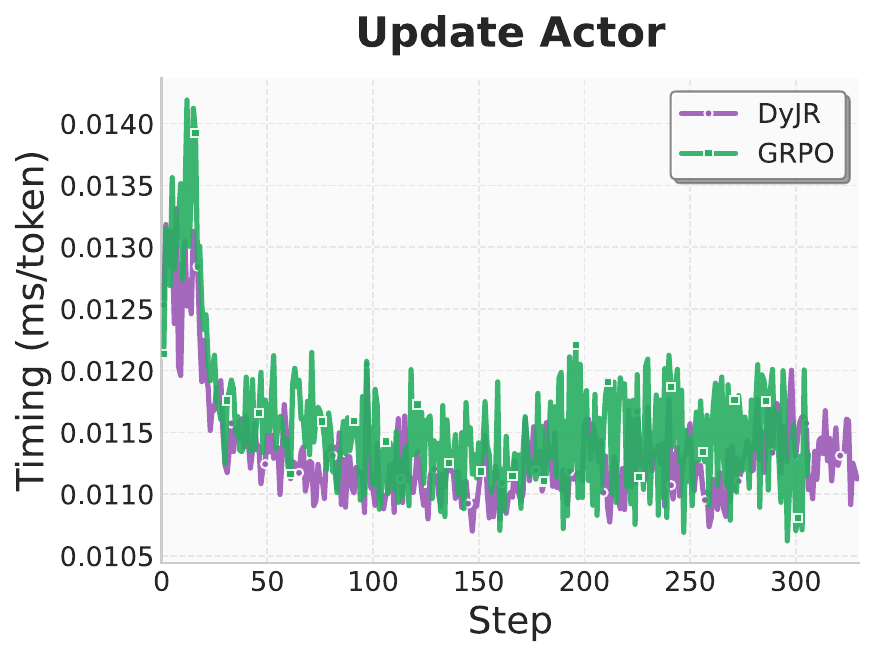}
        \caption{}
        \label{fig:sub_c}
    \end{subfigure}
    
    \caption{(a) Pass@$k$ performance on the \textbf{Beyond AIME} dataset across varying sampling budgets $k$. (b) and (c) Computational latency comparison between GRPO (w/o KL) and DyJR  during the \textit{Rollout} phase and the \textit{Actor Update} phase (encompassing log-probability calculation, forward passes, and gradient updates), respectively.}
    \label{fig:three_graphs}
\end{figure}

On Beyond AIME, DyJR's Pass@$k$ continues improving as the sampling budget increases to 1,024 and exceeds GRPO and the forward-KL variant in the reported curves (Figure~\ref{fig:three_graphs}). This provides a task-level complement to the token diagnostics. Unlike token entropy, Pass@$k$ measures whether a sampling budget uncovers at least one verified solution. Its improvement is therefore relevant to the usefulness of retained alternatives, although the curve alone cannot distinguish better individual samples from less correlated attempts or identify a causal mechanism. The same figure reports rollout and actor-update latency, which remain close to the GRPO baseline in the measured configuration; Appendix~\ref{app:additional-analysis} gives further details.

\begin{table}[t]
\centering
\small
\caption{Measured resident-memory increase (GiB), summed over four H100 GPUs, in an isolated token-ID storage benchmark. Each trajectory has 8,192 \texttt{int32} token IDs. RLEP is tested at maximum per-query capacity; DyJR uses a fixed total capacity. Values include allocation/runtime overhead and exclude other replay fields and training state.}
\label{tab:rebuttal-memory}
\setlength{\tabcolsep}{5pt}
\renewcommand{\arraystretch}{1.18}
\begin{tabular*}{\linewidth}{@{\extracolsep{\fill}}lrrrr@{}}
\toprule
Storage configuration & $Q=10\mathrm{K}$ & $Q=30\mathrm{K}$ & $Q=50\mathrm{K}$ & $Q=100\mathrm{K}$ \\
\midrule
RLEP: 64 trajectories/query & 19.91 & 59.31 & 99.03 & 195.86 \\
\rowcolor{black!4}
DyJR: 1,638 trajectories total & 0.41 & 0.41 & 0.41 & 0.41 \\
\bottomrule
\end{tabular*}
\end{table}
Replay storage depends on the recent admission window: $|\mathcal S_t|\leq\sum_{\tau=t-M}^{t-1}\eta_\tau N$, up to integer rounding and the supply of successful trajectories. At $N=4,096$ and $M=8$, the 20\% warm-up rate allows about 6,554 trajectories; after the window contains only 5\% admission steps, capacity is about 1,638. Thus, the steady-state bound is smaller than the warm-up bound and independent of the total training-set size. Token IDs require integer storage; cached log-probabilities and masks add further memory. Table~\ref{tab:rebuttal-memory} reports an isolated four-GPU token-ID benchmark: DyJR allocation stays constant across query counts, while RLEP's filled per-query capacity scales with them. Appendix~\ref{app:rebuttal-memory} gives the complete measurement protocol. It is not a measurement of complete training memory. The 512 replay rows add forward/backward work at a 1:8 replay-to-online ratio; their practical cost depends on implementation and sequence lengths.

%% file: main_text/discussion.tex
\section{Related Work}

\paragraph{Experience Replay in LLM Training.}
RLVR supplies objective feedback in mathematics and coding~\citep{lambert2025tulu3pushingfrontiers,deepseekmath,qwen2.5,htl}. Building on reasoning systems such as OpenAI o1 and DeepSeek-R1~\citep{o1,guo2025deepseek}, QwQ~\citep{qwq} and Kimi k1.5~\citep{ki1.5} further explore this training paradigm. These developments make rollout reuse particularly relevant for long reasoning trajectories: collecting each response is costly, while the training distribution shifts as the policy improves. Originally developed for off-policy reinforcement learning~\citep{mnih2013playing,lillicrap2015continuous}, experience replay has also been studied in continual learning~\citep{rolnick2019experience} and adapted to LLM training to improve sample efficiency. ReMix~\citep{liang2025squeeze} and RePO~\citep{li2025repo} combine on-policy and off-policy learning, while RLEP~\citep{zhang2025rlep} and Ex-GRPO~\citep{zhan2025exgrpo} focus on reusing high-quality historical trajectories. ARPO~\citep{lu2025arpo} likewise integrates replay with GRPO. These approaches motivate careful treatment of the mismatch between current and historical policies. DyJR uses recent replay as a regularization reference rather than an additional reward-gradient dataset. Our matched controls distinguish the benefit of constructing a populated, recent reference from the incremental benefit of JS-based utilization.

\paragraph{Policy Plasticity Maintenance in LLM RL.}
Rapid concentration during RLVR has motivated work on entropy, exploration, and distributional constraints~\citep{agarwal2025unreasonableeffectivenessentropyminimization,cheng2025reasoningexplorationentropyperspective,tan2025bottom}. Entropy-based techniques~\citep{wang20258020rulehighentropyminority,chen2025seedgrposemanticentropyenhanced} modify the training signal to retain exploratory behavior. SimKO~\citep{peng2025simko} and W-REINFORCE~\citep{zhu2025surprising} differentiate the treatment of successful and unsuccessful samples to counter distribution narrowing. DPH-RL~\citep{li2025choice} studies the $f$-divergence family as an alternative to reverse KL. DyJR complements these directions through a selectively populated, temporally refreshed reference: its regularizer targets recently successful continuations instead of increasing uncertainty indiscriminately. A fixed initial policy supplies the same reference throughout training, whereas temporally local replay tracks successful behaviors reached by intervening policies. The reverse-KL and static-reference controls in Section~\ref{sec:reference-controls} compare these alternatives; their results caution against treating all regularization toward past policies as equivalent. Additional RLVR background appears in Appendix~\ref{app:related-work}.

\section{Conclusion}
We introduced DyJR, a framework that uses experience replay to sustain local policy plasticity during RLVR. A time-sensitive buffer retains recent successful trajectories, and JS regularization uses them as a dynamic anchor alongside the on-policy objective. Experiments on mathematics, Text-to-SQL, and coding demonstrate improvements over the tested baselines. Matched controls show that both adequate buffer admission and the choice of replay objective matter, while a larger early admission rate followed by contraction limits storage without an observed loss in the reported comparison. Rank-token dynamics and semantic-mode analyses support the interpretation that DyJR retains more alternatives around the current policy. These measurements do not establish a universal mechanism, and extending the approach to larger models and more direct behavioral criteria for replay remains an important direction.

%% file: acknowledgments.tex
\section*{Funding and Competing Interests}
This research received no external funding. The authors declare no competing interests.

%% file: appendix/limitations.tex
\section{Limitations and Broader Impacts}
\label{app:limitations}
\paragraph{Scope and uncertainty.}
The experiments cover dense language models from 1.7B to 8B parameters and the specified mathematics, SQL, and coding tasks. They do not establish behavior at substantially larger scales or in settings without reliable verifiable rewards. Three independent runs are available for GRPO, DyJR, RLEP, and Ex-GRPO in the main mathematical comparison, but most auxiliary studies and cross-domain comparisons are single runs. Small performance gaps in those studies should not be interpreted as statistically established differences.

\paragraph{Mechanism and measurement.}
Confidence-based admission and JS utilization have complementary effects. The matched controls do not support attributing all improvement to JS alone. Recency is a proxy for reference relevance, and reward filtering deliberately biases replay toward successful responses. The sampled regularizer is not an exact divergence to an explicitly evaluated mixture. Token statistics and Pass@$k$ are indirect evidence of plasticity; the controlled SQL protocols and LLM-judged mathematical modes address different aspects of diversity. Their finite sample budgets, judge dependence, and inclusion of coherent incorrect strategies prevent conclusions about complete preservation of useful semantic alternatives or a universal causal mechanism.

\paragraph{Selection and hyperparameters.}
The main comparison uses the common step-240 endpoint. Checkpoints were inspected on the evaluation benchmarks without an independent validation split; although the selected endpoint is the same for every compared method and run, this is a limitation of the evaluation protocol. The warm-up boundary of 20 steps was chosen from the observed early transition and is setting-dependent. A matched 10/20/40-step study was not completed, and the available coefficient comparisons are not exhaustive. For a new setting, a short pilot can identify when smoothed loss or entropy leaves the initial rapid-change regime. Some auxiliary experiment details, including the entropy-control coefficient and complete decoding settings for the semantic audits, require more complete archival documentation for exact replication.

\paragraph{Efficiency.}
The buffer's steady-state capacity is smaller than its warm-up capacity. The isolated memory benchmark excludes cached log-probabilities and training state, fills the specified replay capacities, and sums allocations over four devices. Its scaling result should not be equated with complete training memory or the typical occupancy of every replay implementation. End-to-end runtime depends on hardware, sequence lengths, and batching.

\paragraph{Broader impacts.}
More efficient RLVR can reduce the resources needed to adapt reasoning models, while better preservation of alternative solutions may benefit applications with verifiable outcomes. Verification remains task-dependent: a rewarded answer need not imply a faithful reasoning process, and generated code or SQL can still be incorrect or unsuitable for deployment. Stronger generation capabilities can also be misused. The experiments do not establish safety, privacy, fairness, or reliability in deployed systems; independent testing and appropriate execution controls remain necessary in such uses.

%% file: appendix/method_details.tex
\section{Additional Method Details}
\label{app:method-details}
\subsection{GRPO Background}
\label{app:preliminaries}
For a prompt $x$, GRPO samples $G$ responses from the rollout policy $\pi_{\theta_{\mathrm{old}}}$. Let $\mu_x$ and $\sigma_x$ be the group reward mean and standard deviation. The response-level advantage is the group-normalized reward, $\widehat A_i=(r(x,y_i)-\mu_x)/(\sigma_x+\varepsilon)$, with a numerical stabilizer $\varepsilon$. The same advantage applies to the response's tokens. At position $j$, the policy ratio is
\begin{equation}
\rho_{i,j}(\theta)=\frac{\pi_\theta(y_i^{(j)}\mid x,y_i^{(<j)})}{\pi_{\theta_{\mathrm{old}}}(y_i^{(j)}\mid x,y_i^{(<j)})}.
\end{equation}
The standard clipped per-token surrogate is
\begin{equation}
\ell_{i,j}^{\mathrm{GRPO}}=-\min\!\left[\rho_{i,j}\widehat A_i,\ \operatorname{clip}(\rho_{i,j},1-\epsilon_{\mathrm{low}},1+\epsilon_{\mathrm{high}})\widehat A_i\right].
\end{equation}
The training configuration uses token-mean aggregation, $\epsilon_{\mathrm{low}}=0.2$, and $\epsilon_{\mathrm{high}}=0.28$. DyJR adds its replay loss to this online objective. Replay rows do not receive an online GRPO advantage or an off-policy reward-gradient term.

\subsection{Divergence Background and Surrogate Normalization}
\label{app:divergence-background}
For distributions $P$ and $Q$, let $R=(P+Q)/2$. Then
\begin{align}
D_{\mathrm{KL}}(P\|Q)&=\mathbb E_{z\sim P}\log\frac{P(z)}{Q(z)},\\
D_{\mathrm{JS}}(P\|Q)&=\tfrac12D_{\mathrm{KL}}(P\|R)+\tfrac12D_{\mathrm{KL}}(Q\|R).
\label{eq:js-definition}
\end{align}
Conventional JS is symmetric and bounded between zero and $\ln2$. Where the density ratio $u=P/Q$ is defined, direct substitution gives
\begin{equation}
2D_{\mathrm{JS}}(P\|Q)=\mathbb E_{z\sim Q}\left[u\log u-(u+1)\log\frac{u+1}{2}\right].
\label{eq:js-identity}
\end{equation}
This explains the factor of two in the implemented generator and its absorption into $\alpha_{\mathrm{JS}}$. The identity assumes exact sampling from $Q$. DyJR instead uses success-filtered historical trajectories and each trajectory's cached generating-policy probability. The selected historical sample distribution is not the unconditioned generating policy, and its denominator is not an explicitly evaluated mixture density. The resulting token-level sample objective is a JS-motivated surrogate.

Bounded divergence is a property of the full distributional expectation, not an upper bound on individual values of $f(u)$ or their gradients. Likewise, the GRPO clipped surrogate is not a hard constraint on policy distance. Our discussion of a recent policy neighborhood is a motivation tested by recency ablations, not a theoretical guarantee. No trajectory-level importance correction is claimed for this regularization reference.

\subsection{DyJR Algorithm}
\label{app:algorithm}
\begin{algorithm}[htbp]
\caption{Dynamic Jensen-Shannon Replay}
\label{alg:dyjr}
\begin{algorithmic}[1]
\STATE \textbf{Input:} Dataset $\mathcal D$, policy $\pi_\theta$, age limit $M=8$, coefficient $\alpha_{\mathrm{JS}}=0.05$.
\STATE Initialize replay buffer $\mathcal S\leftarrow\varnothing$.
\FOR{training step $t=1,\ldots,T$}
\STATE Freeze rollout parameters $\theta_{\mathrm{old}}\leftarrow\theta$.
\STATE Generate $G=8$ responses for each of 512 prompts; cache sampled-token log-probabilities.
\STATE Compute rewards, group advantages, and prompt confidences $C(x)$.
\STATE Evict entries generated more than $M$ steps ago; current entries have not yet been inserted.
\STATE If $|\mathcal S|\ge512$, draw 512 historical rows uniformly without replacement once; otherwise skip replay.
\STATE Combine online and replay rows with source-type tags and shuffle the pool.
\STATE Make one optimizer epoch with mini-batch size 128: GRPO on online rows; JS on replay rows.
\STATE Keep the selected replay batch fixed throughout the optimizer epoch.
\STATE Set admission rate to 20\% for the first 20 steps and 5\% thereafter.
\STATE Select successful current trajectories by descending $C(x)$, subject to the admission budget.
\STATE \textbf{After optimization}, insert selected trajectories with their original rollout-time log-probabilities and timestamp $t$.
\ENDFOR
\end{algorithmic}
\end{algorithm}
With the full replay batch, the 4,608-row pool yields 36 optimizer steps. This is one pass over the joint pool, not 36 resamplings of replay or repeated epochs. The replay-to-online trajectory ratio is 1:8, and replay occupies one ninth of the combined pool. Individual mini-batches need not preserve that ratio.

%% file: appendix/original_experiments.tex
\section{Additional Related Work}\label{app:related-work}
\paragraph{Reinforcement Learning with Verifiable Rewards} RLVR~\citep{yue2025does, lambert2025tulu3pushingfrontiers} has emerged as a cornerstone for enhancing the reasoning capabilities of LLMs in domains with objective ground truth, such as mathematics and coding~\citep{deepseekmath, guo2025deepseek, qwen2.5, htl}. Following the paradigm established by models like OpenAI o1~\citep{o1} and DeepSeek-R1~\citep{guo2025deepseek}, recent efforts such as QwQ~\citep{qwq} and Kimi k1.5~\citep{ki1.5} have leveraged verifiable feedback to scale test-time compute. While effective, standard RLVR typically relies on on-policy optimization, which often suffers from sample inefficiency and training instability when dealing with long-sequence reasoning trajectories.

\section{Experiment Setups}\label{expset}
\begin{table*}[ht]
    \centering
    \caption{Experimental settings adopted in this work. We employ the AdamW optimizer with $\beta_{1}=0.9, \beta_{2}=0.95$ and a weight decay of $0.1$. The gradient clipping threshold is set to $1.0$ for all models.}
    \resizebox{\textwidth}{!}{%
    \begin{tabular}{lcccccc}
        \toprule
        \textbf{Category} & \textbf{Parameters} & \textbf{GRPO} & \textbf{DAPO} & \textbf{DPH-RL} & \textbf{DyJR} & \textbf{RLEP} \\
        \midrule
        \multirow{7}{*}{Inference} & $G$ (Group Size) & 8 & 8 & 8 & 8 & 8 \\
        & Prompt Batch Size & 512 & 512 & 512 & 512 & 512 \\
        & Temperature & 1.0 & 1.0 & 1.0 & 1.0 & 1.0 \\
        & Top-$p$ / Top-$k$ & 1.0 / -1 & 1.0 / -1 & 1.0 / -1 & 1.0 / -1 & 1.0 / -1 \\
        & Evaluation Temp & 0.7 & 0.7 & 0.7 & 0.7 & 0.7 \\
        & Evaluation Top-$p$ & 0.95 & 0.95 & 0.95 & 0.95 & 0.95 \\
        & Response Length & 6144 & 6144 & 6144 & 6144 & 6144 \\
        \midrule
        \multirow{10}{*}{Training}
        & Clip Ratio High ($\epsilon$) & 0.28 & 0.28 & 0.28 & 0.28 & 0.28 \\
        & Clip Ratio Low ($\epsilon$) & 0.2 & 0.2 & 0.2 & 0.2 & 0.2 \\
        & Learning Rate & $2 \times 10^{-5}$ & $2 \times 10^{-5}$ & $2 \times 10^{-5}$ & $2 \times 10^{-5}$ & $2 \times 10^{-5}$ \\
        & Mini Batch Size & 128 & 128 & 128 & 128 & 128 \\
        & Online Loss Agg. & token-mean & token-mean & token-mean & token-mean & token-mean \\
        & Dynamic Sampling & False & True & True & False & False \\
        & Replay Bsz & - & - & 512 & 512 & 512 \\
        & Correct Threshold & - & - & 6 of 8 & - & - \\
        & Max Age ($M$) & - & - & - & 8 & - \\
        & Regularization Coefficient & - & - & 0.2 & 0.05 & - \\
        \midrule
        \multirow{2}{*}{Hardware} & GPU Count & 32 $\times$ A800 & 32 $\times$ A800 & 32 $\times$ A800 & 32 $\times$ A800 & 32 $\times$ A800 \\
        & Parallel (TP/SP) & 1 / 1 & 1 / 1 & 1 / 1 & 1 / 1 & 1 / 1 \\
        \bottomrule
    \end{tabular}%
    }
    \label{tab:setup}
\end{table*}

We evaluate the following tasks: (1) A \textbf{long-chain mathematical reasoning task} based on the Qwen3-4B-Base~\citep{qwen3} model. For training, we utilize Reinforce-Ada-Hard~\citep{xiong2025reinforce}, a high-difficulty subset of 14k samples filtered from the OpenR1-Math-220k\footnote{https://huggingface.co/datasets/open-r1/OpenR1-Math-220k} dataset, specifically excluding simpler prompts with an average success rate (average reward $>$ 0.375). Evaluation is performed on highly challenging benchmarks, including AIME25~\citep{AIME_AMC}, HMMT25~\citep{balunovic_srimatharena_2025}, BRUMO25~\citep{balunovic_srimatharena_2025}, AMC23, Minerva~\citep{MinervaMath} and Beyond AIME. This task is designed to observe the significant performance leap from a low baseline to high-level proficiency following RL training. (2) A \textbf{SQL generation task} based on Llama-3.1-8B-Instruct~\citep{meta2024llama3}, which focuses on moderate-difficulty reasoning typically within a 1k token limit. The model is trained on the BIRD~\citep{li2024bird} dataset and evaluated across both BIRD and Spider~\citep{yu2018spider} datasets to test cross-domain performance.

For coding, we train Qwen2.5-7B-Instruct on CodeContests problems of difficulty at most 2 and evaluate LiveCodeBench-V2 and a held-out CodeContests split, reporting Pass@16 under unit-test verification.

In this work, we employ multiple experimental setups to validate the effectiveness of \textbf{DyJR} in comparison with baseline methods; Table~\ref{tab:setup} summarizes the main mathematical training configuration. SQL uses the shorter responses described above; the isolated storage study has its own hardware and length settings in Appendix~\ref{app:rebuttal-memory}. Specifically, \textit{Replay Bsz} denotes the historical replay batch used in an update, not the number of newly admitted trajectories. New admission is controlled separately by the 20\%/5\% schedule. For \textbf{DPH-RL}, following the experimental protocol in their original work, we incorporate samples that yield 6 correct responses out of 8 sampled paths into the regularization term. Regarding \textbf{RLEP}, we admit two correct solution trajectories per query per update, with up to 64 retained trajectories per query. These buffered trajectories participate in the replay baseline's policy updates, in contrast with DyJR's regularization-only utilization. For Pass@1, we employ greedy decoding (i.e., temperature $T=0$). For Mean@256, we generate 256 samples with a temperature of $T=0.7$ and calculate the average accuracy across all samples.

\paragraph{ExGRPO.}(Experiential Group Relative Policy Optimization) ~\citep{zhan2025exgrpo} extends GRPO with an experience replay framework that integrates \emph{experience management} and \emph{mixed optimization}. During training, ExGRPO maintains an experience pool that maps each prompt to a set of previously successful solution trajectories, and categorizes prompts into difficulty buckets according to their most recent rollout accuracy. To prevent overfitting to trivial cases, prompts whose rollouts are entirely correct are moved into a \emph{retired set} and excluded from subsequent replay.

At each update step, ExGRPO applies a replay ratio $\rho$ to partition each mini-batch into on-policy samples and replayed experiences. Difficulty buckets are sampled using a Gaussian weighting centered at $0.5$, thereby emphasizing prompts of intermediate difficulty. For each selected prompt, ExGRPO further selects the successful trajectory with the lowest entropy under the current policy, encouraging stable exploitation while mitigating excessive plasticity collapse.

The additional hyperparameters introduced by ExGRPO are summarized in Table~\ref{tab:exgrpo_config}, while all other training configurations follow the GRPO baseline in the Table \ref{tab:setup}.

\begin{table}[t]
\centering
\small
\caption{ExGRPO-specific hyperparameters for experience replay.}
\label{tab:exgrpo_config}
\begin{tabular}{lll}
\toprule
\textbf{Parameter} & \textbf{Value} & \textbf{Description} \\
\midrule
Replay ratio ($\rho$) & 0.5 & Fraction of replay samples in each mini-batch \\
Delayed replay threshold & 0.35 & Minimum batch Pass@1 required to activate experience replay \\
Difficulty bucket range & $[0, 7]$ & Bounds for difficulty-based bucketing ($k/K$) \\
Replay metric & Entropy & Criterion for evaluating trajectory quality \\
Trajectory selection mode & Argmin & Select the lowest-entropy successful trajectory per prompt \\
\bottomrule
\end{tabular}
\end{table}

\paragraph{Training budget and evaluation.}
Main long-chain runs use 32 A800 GPUs and take approximately 20 hours each, or approximately 640 GPU-hours per run. Checkpoints are evaluated every 80 steps, and step 240 is the reported endpoint for every method and run in the main comparison. No separate validation set is used. The reported runtime does not provide a complete aggregate accounting of all pilots, auxiliary studies, or failed runs.

\section{Admission Schedule and Replay Controls}
\label{app:admission-schedule}
\begin{table*}[h]
\centering
\setlength{\tabcolsep}{4pt}
\renewcommand{\arraystretch}{1.05}

\caption{Performance of DyJR on various mathematical benchmarks. $init$ denotes the percentage of rollout samples retained per step during the initial 20 steps, while $\eta$ represents the retention rate thereafter. The maximum age of the FIFO buffer is consistently set to 8 steps.}
\label{tab:math_results_dyjr}

\resizebox{0.95\linewidth}{!}{%
\begin{tabular}{lccccccc}
\toprule
\textbf{Model} & \textbf{AIME25} & \textbf{AMC23} & \textbf{Beyond AIME} & \textbf{BRUMO25} & \textbf{HMMT25} & \textbf{Minerva}* & \textbf{Avg} \\
\midrule

DyJR ($init=20\%, \eta=20\%$) & 22.9 & \textbf{72.5} & 12.5 & \textbf{33.8} & 12.6 & \textbf{51.0} & \textbf{34.2} \\

DyJR ($init=20\%, \eta=5\%$) & \textbf{23.1} & 72.0 & \textbf{12.7} & 33.4 & \textbf{12.7} & 50.4 & 34.1 \\

DyJR ($init=10\%, \eta=5\%$) & 22.5 & 69.2 & 11.4 & 31.5 & 11.5 & 48.2 & 32.4 \\

DyJR ($init=5\%, \eta=5\%$) & 21.8 & 68.5 & 10.1 & 29.2 & 10.3 & 47.8 & 31.3 \\

\bottomrule
\end{tabular}%
}
\end{table*}

Table~\ref{tab:math_results_dyjr} compares admission rates under the same age constraint. Constant 20\% admission and the 20\%$\rightarrow$5\% schedule obtain 34.2 and 34.1. The result supports storage-efficient contraction after adequate early population, without establishing that late trajectories are intrinsically less valuable. Lower initial admission yields 32.4 at 10\% and 31.3 at 5\%. These single-run comparisons do not supply confidence intervals for the small 34.2--34.1 difference.
\section{Replay Selection and Sparse Rewards}
\label{app:selection-sparse-rewards}

Filtering replay samples by reward introduces selection bias. DyJR does not aim to estimate the full rollout distribution unbiasedly; instead, it constructs a high-quality reference for plasticity regularization. A stricter rule that admits successful trajectories only from fully correct prompt groups can cause buffer starvation in harder settings. The confidence-stratified admission rule improves the tradeoff between replay quality and replay availability.

Table~\ref{tab:selection_bias_sparse} compares perfect-only replay with confidence-based replay. Increasing $M$ from 8 to 32 raises the perfect-only result from 30.4 to 31.3, but reduces the confidence-based result from 34.1 to 31.7. More storage can alleviate scarcity under a restrictive rule, whereas retaining older samples can hurt when fresh anchors are already available. The matched admission--objective comparison appears in Table~\ref{tab:admission-main}.

\begin{table}[h]
\centering
\small
\caption{Replay selection and staleness. Confidence-based selection reduces buffer starvation, while overly stale replay remains harmful.}
\label{tab:selection_bias_sparse}
\begin{tabular}{lcc}
\toprule
\textbf{Replay Selection} & \textbf{Max Age $M$} & \textbf{Avg} \\
\midrule
Perfect-only & 8 & 30.4 \\
Perfect-only & 32 & 31.3 \\
Confidence-based & 8 & \textbf{34.1} \\
Confidence-based & 32 & 31.7 \\
\bottomrule
\end{tabular}
\end{table}

We also evaluate a harder sparse-reward setting with Qwen3-1.7B-Base, where the initial reward is much lower and final reward remains modest. Table~\ref{tab:sparse_reward_17b} shows that DyJR still improves over GRPO, suggesting that the method does not rely on a particularly strong initial policy.

\begin{table}[h]
\centering
\small
\caption{Sparse-reward setting with Qwen3-1.7B-Base.}
\label{tab:sparse_reward_17b}
\begin{tabular}{lc}
\toprule
\textbf{Method} & \textbf{Avg} \\
\midrule
GRPO (1.7B, sparse) & 17.3 \\
DyJR (1.7B, sparse) & \textbf{19.5} \\
\bottomrule
\end{tabular}
\end{table}

\section{Alternative Regularizers and Off-Policy Corrections}
\label{app:alternative-regularizers}

We further compare DyJR with common alternatives under the same compute budget. The evaluated alternatives include static-reference KL, a dynamic entropy bonus, and direct off-policy replay with importance ratios. These single-run results compare the specified configurations rather than all possible variants of each regularizer. Table~\ref{tab:alternative_regularizers} summarizes the results.

\begin{table}[h]
\centering
\small
\caption{Comparison with alternative regularization and off-policy correction strategies under the same compute budget.}
\label{tab:alternative_regularizers}
\begin{tabular}{lc}
\toprule
\textbf{Method} & \textbf{Avg} \\
\midrule
KL-to-ref (static reference) & 30.8 \\
Dynamic entropy bonus & 30.3 \\
Replay + off-policy PG w/ IS ratio & 32.6 \\
DyJR & \textbf{34.1} \\
\bottomrule
\end{tabular}
\end{table}

%% file: appendix/rebuttal_additions.tex
\section{Additional Evaluation of Replay, Diversity, and Memory}
\label{app:rebuttal-evaluation}

This section evaluates the robustness of replay-baseline comparisons, separates replay admission from replay utilization, and examines diversity beyond token-level summaries. We also report an isolated storage benchmark. These experiments assess complementary aspects of DyJR; none alone establishes a complete causal account of local policy plasticity.

\subsection{Repeated Runs of Replay Baselines}
\label{app:rebuttal-repeated-runs}

We evaluate Ex-GRPO and RLEP with three independent runs each, including the originally reported run, in the same VERL training stack used for the main comparisons. GRPO and DyJR also use three independent runs. Method-specific hyperparameters follow the respective implementations. Checkpoints are evaluated every 80 training steps on the same six benchmarks, without a separate validation set. For every method and run in this comparison, the highest six-benchmark average occurs at the final evaluated checkpoint, step 240. Thus, the reported comparison is identical to reporting step 240 uniformly, although benchmark-based checkpoint selection can introduce optimistic bias in general.

Table~\ref{tab:repeated-run-uncertainty} reports the means and sample standard deviations for all four repeated-run methods; Table~\ref{tab:rebuttal-replay-seeds} gives the individual Ex-GRPO and RLEP runs. Ex-GRPO obtains $32.6$ and RLEP obtains $31.8$ on the six-benchmark average, compared with $34.1$ for DyJR. The observed mean differences are 1.5 and 2.3 percentage points, respectively. We report these as repeated-run comparisons rather than formal significance tests; three runs provide only a limited estimate of training variability.

\begin{table}[htbp]
\centering
\small
\caption{Variability across three independent training runs. Entries are reported means $\pm$ sample standard deviations, in percentage points. Minerva uses Mean@16; the other benchmarks use Mean@256. Avg is the six-benchmark average. The aggregate standard deviation for GRPO was not reported, so only its mean is shown.}
\label{tab:repeated-run-uncertainty}
\setlength{\tabcolsep}{5pt}
\renewcommand{\arraystretch}{1.18}
\begin{tabular*}{\linewidth}{@{\extracolsep{\fill}}lrrrr@{}}
\toprule
Benchmark & GRPO & RLEP & Ex-GRPO & DyJR \\
\midrule
AIME25      & $19.1 \pm 1.8$ & $22.6 \pm 0.5$ & $22.2 \pm 0.4$ & $23.1 \pm 1.6$ \\
AMC23       & $64.6 \pm 1.5$ & $70.9 \pm 1.7$ & $69.9 \pm 0.7$ & $72.0 \pm 1.1$ \\
Beyond AIME & $10.7 \pm 0.9$ & $10.9 \pm 0.3$ & $11.4 \pm 1.7$ & $12.7 \pm 0.7$ \\
BRUMO25     & $29.6 \pm 1.6$ & $28.7 \pm 0.8$ & $33.0 \pm 0.2$ & $33.4 \pm 1.4$ \\
HMMT25      & $ 9.8 \pm 0.8$ & $10.7 \pm 1.6$ & $10.0 \pm 1.3$ & $12.7 \pm 0.8$ \\
Minerva     & $45.2 \pm 1.4$ & $47.3 \pm 1.5$ & $49.2 \pm 0.9$ & $50.4 \pm 2.6$ \\
\midrule
Avg         & $29.8\phantom{{} \pm 0.0}$ & $31.8 \pm 0.2$ & $32.6 \pm 0.2$ & $\textbf{34.1} \pm 0.3$ \\
\bottomrule
\end{tabular*}
\end{table}

\begin{table}[htbp]
\centering
\small
\caption{Individual runs of replay baselines. Values are percentages; the final column is the average over six benchmarks. Run 1 is the originally reported run. Reported rounded values are preserved; their three-run summaries appear in Table~\ref{tab:repeated-run-uncertainty}.}
\label{tab:rebuttal-replay-seeds}
\setlength{\tabcolsep}{3pt}
\renewcommand{\arraystretch}{1.15}
\begin{tabular*}{\linewidth}{@{\extracolsep{\fill}}lcrrrrrrr@{}}
\toprule
Method & Run & AIME25 & AMC23 & \shortstack{Beyond\\AIME} & BRUMO25 & HMMT25 & Minerva* & Avg \\
\midrule
Ex-GRPO & 1 & 21.7 & 70.2 & 11.6 & 32.8 & 10.4 & 50.1 & 32.8 \\
        & 2 & 22.5 & 70.3 & 12.9 & 33.0 & 8.5 & 48.4 & 32.6 \\
        & 3 & 22.4 & 69.1 & 9.6 & 33.2 & 11.1 & 49.0 & 32.4 \\
\midrule
RLEP    & 1 & 23.0 & 68.9 & 10.6 & 28.9 & 10.8 & 48.4 & 31.8 \\
        & 2 & 22.7 & 72.2 & 11.1 & 29.3 & 9.1 & 47.8 & 32.0 \\
        & 3 & 22.0 & 71.5 & 11.1 & 27.8 & 12.3 & 45.6 & 31.7 \\
\bottomrule
\end{tabular*}
\end{table}

\subsection{Separating Replay Admission from Utilization}
\label{app:rebuttal-admission-utilization}

Confidence-based admission changes which trajectories enter the buffer, whereas the replay objective determines how the available trajectories affect optimization. To separate these factors, we compare perfect-only and confidence-based admission under both JS regularization and off-policy policy-gradient replay with importance sampling (PG+IS). All four settings use $M=8$ and matched buffer configuration and training budget. The confidence-based PG+IS setting uses the same default admission rule as DyJR.

\begin{table}[htbp]
\centering
\small
\caption{Matched admission--utilization comparison. All settings use $M=8$ and the same replay-buffer configuration and training budget. Avg is the six-benchmark average accuracy (\%).}
\label{tab:rebuttal-admission-utilization}
\begin{tabular}{llr}
\toprule
Admission & Replay utilization & Avg \\
\midrule
Perfect-only & JS regularization & 30.4 \\
Confidence-based & JS regularization (DyJR) & \textbf{34.1} \\
Perfect-only & Off-policy PG+IS & 30.1 \\
Confidence-based & Off-policy PG+IS & 32.6 \\
\bottomrule
\end{tabular}
\end{table}

Changing from perfect-only to confidence-based admission improves the average from 30.4 to 34.1 with JS and from 30.1 to 32.6 with PG+IS. This is consistent with the importance of avoiding buffer starvation when fully correct trajectories are sparse. Once confidence-based admission is fixed, JS improves the average by a further 1.5 points over PG+IS. These results support complementary roles: admission provides sufficient fresh anchors, and the JS objective makes more effective use of those anchors in this setting. They do not attribute the entire gain over GRPO to JS alone.

\subsection{Standard Token-Level Entropy Regularization}
\label{app:rebuttal-entropy}

We add a standard token-level entropy bonus to GRPO under the same training setup:
\begin{equation}
\mathcal{L}_{\mathrm{entropy}}
=\mathcal{L}_{\mathrm{GRPO}}
-\beta\,\mathbb{E}_{t}\!\left[H\!\left(\pi_\theta(\cdot\mid s_t)\right)\right],
\label{eq:rebuttal-entropy}
\end{equation}
where $\beta$ is the entropy coefficient. Table~\ref{tab:rebuttal-entropy} shows essentially unchanged average accuracy for this configuration: 29.7, compared with 29.8 for GRPO and 34.1 for DyJR. The entropy bonus encourages uncertainty across response-token distributions, whereas DyJR conditions its preservation signal on recently accepted replay trajectories. This comparison supports the value of a targeted replay anchor in the evaluated configuration; it is not an exhaustive study of entropy coefficients or schedules.

\begin{table}[htbp]
\centering
\small
\caption{Standard token-level entropy regularization. Avg is the six-benchmark average accuracy (\%).}
\label{tab:rebuttal-entropy}
\begin{tabular}{lr}
\toprule
Method & Avg \\
\midrule
GRPO & 29.8 \\
GRPO + standard entropy regularization & 29.7 \\
DyJR & \textbf{34.1} \\
\bottomrule
\end{tabular}
\end{table}

\subsection{Controlled Protocol Diversity in NL2SQL}
\label{app:rebuttal-sql-modes}

Token-level entropy cannot distinguish alternative reasoning protocols from lexical variation. We therefore conduct a controlled NL2SQL experiment with five pre-constructed protocol families used for supervised fine-tuning. Following the multi-path construction in CHASE-SQL~\citep{pourreza2024chasesql}, these comprise divide-and-conquer reasoning, query-plan-first reasoning, direct step-by-step reasoning, and trajectories distilled from DeepSeek-R1 and GPT-4o. Each family has a distinct identifying prefix (Table~\ref{tab:rebuttal-sql-protocols}); the prefix is a label for the constructed family, rather than an independent definition of semantic difference.

\begin{table}[htbp]
\centering
\small
\caption{Five protocol families used in the controlled NL2SQL study. Distinct prefixes provide deterministic family labels.}
\label{tab:rebuttal-sql-protocols}
\begin{tabular}{p{0.36\linewidth}p{0.55\linewidth}}
\toprule
Protocol family & Identifying response prefix \\
\midrule
Divide-and-conquer & ``Divide and Conquer \ldots'' \\
Query-plan-first & ``Query Plan \ldots'' \\
Direct step-by-step & ``Let's break down and think step by step \ldots'' \\
DeepSeek-R1-distilled & ``Okay, to answer the question \ldots'' \\
GPT-4o-distilled & ``The original question can \ldots'' \\
\bottomrule
\end{tabular}
\end{table}

For each test question, we draw 32 responses and count the distinct protocol labels observed. Before RL, approximately 80\% of questions exhibit at least five labels. After GRPO, 95\% exhibit only one. After DyJR, 32\% exhibit two, 30\% three, and 27\% four, totaling 89\% with at least two observed protocol families (Table~\ref{tab:rebuttal-sql-modes}). This controlled result indicates that DyJR maintains a broader set of constructed solution protocols under the same sample budget. It does not establish that every retained response is correct, or that a protocol prefix by itself verifies the semantics of an unconstrained derivation.

\begin{table}[htbp]
\centering
\small
\caption{Observed protocol diversity in NL2SQL with 32 samples per question. Percentages describe questions, not individual responses.}
\label{tab:rebuttal-sql-modes}
\begin{tabular}{lp{0.67\linewidth}}
\toprule
Model & Observed protocol counts \\
\midrule
SFT base & Approximately 80\% of questions exhibit at least five protocols. \\
GRPO & 95\% of questions exhibit one protocol. \\
DyJR & 32\% exhibit two; 30\% three; 27\% four; 89\% exhibit at least two. \\
\bottomrule
\end{tabular}
\end{table}

\subsection{Mathematical Strategies from Complete Derivations}
\label{app:rebuttal-math-modes}

To examine mathematical reasoning beyond prefixes, we fix 50 problems and sample 16 responses per problem from each of Qwen3-4B-Base, GRPO, and DyJR, giving 800 responses per model. For each model--problem pair, GPT-5.6-sol receives the problem, all 16 complete responses, and fixed correctness labels. The judge identifies the mathematical method executed by each response and clusters responses that share the same core strategy.

Coherent but incorrect attempts are retained. Differences restricted to wording, calculation order, arithmetic or algebraic errors, or minor reparameterizations are merged when the underlying strategy is unchanged. Responses without an identifiable strategy are excluded. Consequently, the reported modes describe identifiable mathematical strategies and are not restricted to strategies containing a correct response.

DyJR exhibits at least two modes on 58\% of problems, compared with 12\% for GRPO, and its mean mode count is 2.2 rather than 1.3. Both measures remain below the base model's 78\% and 2.6, respectively. Thus, the result supports greater observed strategy diversity after DyJR than after GRPO on this fixed problem set; it does not imply complete preservation of the base model's alternatives.

For example, AMC 2023 Problem 17 asks for the number of digits in $8^5\cdot5^{10}\cdot15^5$. The base model exhibits both exact prime-factor pairing, yielding $243\cdot10^{15}$ before counting digits, and the logarithmic rule $\lfloor\log_{10}N\rfloor+1$. On AIME 2024 item 24 in the evaluation set, the identified alternatives include logarithmic linearization followed by solving a linear system, and direct substitution after converting the equations to exponential form.

This analysis uses an LLM judge and a small fixed set of problems. Clustering can be sensitive to judge decisions and sampling coverage, and diversity among all identifiable strategies differs from diversity among correct solutions. Moreover, comparing GRPO with the complete DyJR method does not isolate the contribution of JS from that of admission and buffer construction. We therefore treat these results as complementary empirical evidence for the plasticity interpretation, alongside the matched objective ablation and sampling-budget analyses.

\subsection{Isolated Replay-Storage Scaling}
\label{app:rebuttal-memory}

We isolate storage scaling by materializing replay buffers on four NVIDIA H100 80GB GPUs. Every trajectory contains 8,192 tokens represented solely by \texttt{int32} token IDs; stored log probabilities, masks, other metadata, model parameters, and optimizer states are excluded. RLEP is evaluated at its specified maximum capacity of 64 trajectories per training query. Although it collects two trajectories per query at an update, that collection rate is distinct from its maximum retained capacity. The DyJR scenario fixes the total capacity to 1,638 trajectories, independently of the number of training queries $Q$.

The metric is the net increase in resident GPU memory reported by \texttt{nvidia-smi}, summed across the four devices while the complete buffers are live. It includes allocation/runtime overhead beyond the tensor payload, and is not the total memory consumed by a training run. For example, the DyJR token-ID tensors occupy approximately 0.050\,GiB, whereas the measured resident-memory increase is 0.41\,GiB.

The measured RLEP footprint increases with the number of queries because the benchmark fills a separate buffer for each query. The fixed-capacity DyJR footprint remains constant across the tested query counts. These measurements establish the storage-scaling difference under the stated capacity and representation choices; they do not imply that every RLEP run reaches full capacity or that a complete DyJR replay cache contains only token IDs.

\subsection{Interpretation of Post-Warm-Up Contraction}
\label{app:rebuttal-contraction}

Constant 20\% admission obtains an average of 34.2, compared with 34.1 for the $20\%\!\rightarrow\!5\%$ schedule. This comparison supports reducing replay admission and capacity after warm-up without an observed performance cost in this setting. It does not establish that later trajectories are intrinsically less useful. Conversely, the reduced-initial-admission ablations indicate the importance of adequate early buffer population. The evidence therefore favors sufficient early anchor availability followed by storage-efficient contraction, rather than a general ranking of early versus late data quality.

%% file: appendix/additional_analysis.tex
\section{Additional Sampling-Budget and Timing Analysis}
\label{app:additional-analysis}
We evaluate Beyond AIME as the sampling budget increases to 1,024 responses per problem. The Pass@$k$ curves in Figure~\ref{fig:three_graphs} place JS-based DyJR above GRPO and the forward-KL variant throughout the reported range. The rollout and actor-update panels show similar measured latency for GRPO and DyJR in this configuration; they are not a universal bound on runtime overhead. Actor updates include log-probability calculation, forward passes, and gradient updates. The replay regularizer is $\mathcal L_{\mathrm{JS}}$, not an SFT objective.

%% file: checklist.tex
\section*{NeurIPS Paper Checklist}

\begin{enumerate}

\item {\bf Claims}
    \item[] Question: Do the main claims made in the abstract and introduction accurately reflect the paper's contributions and scope?
    \item[] Answer: \answerYes{}.
    \item[] Justification: The abstract and introduction describe the empirical scope and the complementary admission and utilization findings; limitations of the mechanism evidence are discussed in Section~\ref{sec:analysis} and Appendix~\ref{app:limitations}.
    

\item {\bf Limitations}
    \item[] Question: Does the paper discuss the limitations of the work performed by the authors?
    \item[] Answer: \answerYes{}.
    \item[] Justification: Appendix~\ref{app:limitations} discusses model scale, sampling and judge limitations, checkpoint selection, uncertainty, hyperparameters, and efficiency measurement.
    

\item {\bf Theory assumptions and proofs}
    \item[] Question: For each theoretical result, does the paper provide the full set of assumptions and a complete (and correct) proof?
    \item[] Answer: \answerNA{}.
    \item[] Justification: The paper is empirical and does not claim new formal theoretical guarantees. Standard objectives and the algebraic normalization of the regularizer are specified in Section~\ref{sec:method} and Appendix~\ref{app:method-details}.
    

    \item {\bf Experimental result reproducibility}
    \item[] Question: Does the paper fully disclose all the information needed to reproduce the main experimental results of the paper to the extent that it affects the main claims and/or conclusions of the paper (regardless of whether the code and data are provided or not)?
    \item[] Answer: \answerNo{}.
    \item[] Justification: The main training configuration, algorithm, update schedule, evaluation metrics, and checkpoint protocol are supplied. Exact replication of some auxiliary semantic and entropy studies still requires parameters and archived records noted in Appendix~\ref{app:limitations}.
    

\item {\bf Open access to data and code}
    \item[] Question: Does the paper provide open access to the data and code, with sufficient instructions to faithfully reproduce the main experimental results, as described in supplemental material?
    \item[] Answer: \answerNo{}.
    \item[] Justification: The source datasets and model families are identified, but this version does not supply a verified public URL for the exact training implementation and complete auxiliary-study records.
    

\item {\bf Experimental setting/details}
    \item[] Question: Does the paper specify all the training and test details (e.g., data splits, hyperparameters, how they were chosen, type of optimizer) necessary to understand the results?
    \item[] Answer: \answerNo{}.
    \item[] Justification: Section~\ref{sec:experiments} and Appendix~\ref{expset} specify the main settings. Some auxiliary-study details, including the entropy coefficient and complete semantic-audit decoding settings, are not fully documented.
    

\item {\bf Experiment statistical significance}
    \item[] Question: Does the paper report error bars suitably and correctly defined or other appropriate information about the statistical significance of the experiments?
    \item[] Answer: \answerYes{}.
    \item[] Justification: Three independent runs are reported for GRPO, DyJR, RLEP, and Ex-GRPO in the main mathematical comparison. Table~\ref{tab:repeated-run-uncertainty} in the appendix reports sample standard deviations across runs; single-run studies are explicitly identified, and no formal significance test is claimed.
    

\item {\bf Experiments compute resources}
    \item[] Question: For each experiment, does the paper provide sufficient information on the computer resources (type of compute workers, memory, time of execution) needed to reproduce the experiments?
    \item[] Answer: \answerNo{}.
    \item[] Justification: Appendix~\ref{expset} reports the main hardware and approximate per-run cost, and Appendix~\ref{app:rebuttal-memory} specifies the isolated memory benchmark. A complete total-project accounting, including pilots and failed runs, is unavailable.
    
    
\item {\bf Code of ethics}
    \item[] Question: Does the research conducted in the paper conform, in every respect, with the NeurIPS Code of Ethics \url{https://neurips.cc/public/EthicsGuidelines}?
    \item[] Answer: \answerYes{}.
    \item[] Justification: The work uses existing research datasets and models, reports empirical limitations, and includes no human-subject study. The described research does not require a stated deviation from the NeurIPS Code of Ethics.
    

\item {\bf Broader impacts}
    \item[] Question: Does the paper discuss both potential positive societal impacts and negative societal impacts of the work performed?
    \item[] Answer: \answerYes{}.
    \item[] Justification: Appendix~\ref{app:limitations} discusses resource-efficiency benefits, verification and reliability limits, and the need to assess risks before deployment.
    
    
\item {\bf Safeguards}
    \item[] Question: Does the paper describe safeguards that have been put in place for responsible release of data or models that have a high risk for misuse (e.g., pre-trained language models, image generators, or scraped datasets)?
    \item[] Answer: \answerNA{}.
    \item[] Justification: This manuscript does not accompany a new high-risk model or scraped-dataset release requiring additional release safeguards.
    

\item {\bf Licenses for existing assets}
    \item[] Question: Are the creators or original owners of assets (e.g., code, data, models), used in the paper, properly credited and are the license and terms of use explicitly mentioned and properly respected?
    \item[] Answer: \answerNo{}.
    \item[] Justification: Model and dataset sources are credited in Section~\ref{sec:experiments} and Appendix~\ref{expset}. Version-specific licenses and terms are not fully documented for all assets.
    

\item {\bf New assets}
    \item[] Question: Are new assets introduced in the paper well documented and is the documentation provided alongside the assets?
    \item[] Answer: \answerNA{}.
    \item[] Justification: This manuscript does not introduce a new dataset or model release. The method and experiments are described in the paper and appendices.
    

\item {\bf Crowdsourcing and research with human subjects}
    \item[] Question: For crowdsourcing experiments and research with human subjects, does the paper include the full text of instructions given to participants and screenshots, if applicable, as well as details about compensation (if any)? 
    \item[] Answer: \answerNA{}.
    \item[] Justification: The experiments do not involve crowdsourcing or research with human subjects.
    

\item {\bf Institutional review board (IRB) approvals or equivalent for research with human subjects}
    \item[] Question: Does the paper describe potential risks incurred by study participants, whether such risks were disclosed to the subjects, and whether Institutional Review Board (IRB) approvals (or an equivalent approval/review based on the requirements of your country or institution) were obtained?
    \item[] Answer: \answerNA{}.
    \item[] Justification: The experiments do not involve human subjects, so an IRB approval is not applicable.
    

\item {\bf Declaration of LLM usage}
    \item[] Question: Does the paper describe the usage of LLMs if it is an important, original, or non-standard component of the core methods in this research? Note that if the LLM is used only for writing, editing, or formatting purposes and does \emph{not} impact the core methodology, scientific rigor, or originality of the research, declaration is not required.
    \item[] Answer: \answerYes{}.
    \item[] Justification: Appendices~\ref{app:rebuttal-sql-modes} and~\ref{app:rebuttal-math-modes} identify the DeepSeek-R1/GPT-4o-distilled SQL protocols and GPT-5.6-sol mathematical-strategy judge. These empirical roles go beyond writing assistance.
    

\end{enumerate}